%% file: main.tex
\documentclass{article}

\input{math_commands_icml.tex}

\usepackage{microtype}
\usepackage{graphicx}
\usepackage{subfigure}
\usepackage{booktabs} 

\usepackage{hyperref}

\usepackage[accepted]{icml2023}

\usepackage{amsmath}
\usepackage{amssymb}
\usepackage{mathtools}
\usepackage{amsthm}

\usepackage[capitalize,noabbrev]{cleveref}

\theoremstyle{plain}
\newtheorem{theorem}{Theorem}[section]
\newtheorem{proposition}[theorem]{Proposition}
\newtheorem{lemma}[theorem]{Lemma}

\theoremstyle{definition}
\newtheorem{definition}[theorem]{Definition}

\theoremstyle{remark}

\usepackage[textsize=tiny]{todonotes}

\icmltitlerunning{On the Equivalence of Consistency-Type Models}

\begin{document}

\twocolumn[
\icmltitle{On the Equivalence of Consistency-Type Models: Consistency Models, Consistent Diffusion Models, and Fokker-Planck Regularization}



\icmlsetsymbol{equal}{*}

\begin{icmlauthorlist}
\icmlauthor{Chieh-Hsin Lai}{sonyai}
\icmlauthor{Yuhta Takida}{sonyai}
\icmlauthor{Toshimitsu Uesaka}{sonyai}
\icmlauthor{Naoki Murata}{sonyai}
\icmlauthor{Yuki Mitsufuji}{sonyai,sony}
\icmlauthor{Stefano Ermon}{sch}
\end{icmlauthorlist}

\icmlaffiliation{sonyai}{Sony AI, Tokyo, Japan}
\icmlaffiliation{sony}{Sony Group Corporation, Tokyo, Japan}
\icmlaffiliation{sch}{Department of Computer Science, Stanford University, Stanford, CA, USA}

\icmlcorrespondingauthor{Chieh-Hsin Lai}{Chieh-hsin.lai@sony.com}

\icmlkeywords{diffusion models, score-based diffusion models, fokker-planck equation, consistency}

\vskip 0.3in
]



\printAffiliationsAndNotice{}  

\begin{abstract}
The emergence of various notions of ``consistency'' in diffusion models has garnered considerable attention and helped achieve improved sample quality, 
likelihood estimation, and accelerated sampling. 
Although similar concepts have been proposed in the literature, the precise relationships among them remain unclear. In this study, we establish theoretical connections between three recent
``consistency'' notions  designed to enhance diffusion models for distinct objectives. Our insights offer the potential for a more comprehensive and encompassing framework for consistency-type models.
\end{abstract}

\section{Introduction}
\label{sec:intro}
Score-based generative models~\citep{song2019generative,song2020score,song2020sliced,boffi2022probability}, commonly referred to as diffusion models~\citep{sohl2015deep,ho2020denoising}, have significantly advanced  photorealistic image generation~\citep{saharia2022photorealistic,rombach2022high,kim2022refining} and found applications in diverse domains such as media editing and restoration~\citep{meng2021sdedit,kawar2022denoising,saito2022unsupervised,murata2023gibbsddrm}. 


Underlying score-based generative model is a (stochastic) differential equation that describes the process of transforming data to noise and vice-versa, which is approximated using a neural (score) network learned from data. Because of the mathematical structure afforded by the underlying differential equation, this neural network needs to satisfy certain consistency properties.
Various such notions of consistency have been recently introduced and shown to
enhance sample quality~\citep{daras2023consistent}, accelerate sampling speed~\citep{song2023consistency}, and improve likelihood estimation~\citep{Lai2022ImprovingSD}. We introduce the term \emph{consistency-type model} to encompass and unify these various concepts. It refers to a (diffusion) model that is explicitly designed to align with the underlying trajectory defined by an ordinary differential equation (ODE), stochastic differential equation (SDE), or partial differential equation (PDE).
In this study, we aim to provide a theoretical investigation into the relationships between these three consistency-type models. Under certain mild assumptions, we will rigorously establish the 
equivalence of these independently developed concepts. 

\section{Background}
\label{sec:background}

\citet{song2020score} introduced a stochastic differential equation (SDE) framework that unifies the concepts of denoising score matching~\citep{song2019generative} and diffusion models~\citep{sohl2015deep,ho2020denoising}  in continuous time. Especially\footnote{Within this study, our primary emphasis is placed on the variance exploding (VE) SDE~\citep{song2020score}. VE SDE entails a process devoid of drift in the forward SDE formulation. Our discussion can be extended to a broader range of forward SDEs.}, the process $\{\bm{x}(t)\}_{t\in[0,T]}$ of adding Gaussian noise
\begin{equation*}
\bm{x}(t)\sim q_t \quad \text{where} \quad q_t(\bm{x}) = q_{\text{data}}(\bm{x})*\mathcal{N}(\bm{x}; \bm{0},\sigma^2(t)\bm{I}) 
\end{equation*}
is driven by the following forward  SDE
\begin{equation}\label{eq:sde_forward}
    d\bm{x}(t) =  g(t) d\bm{w}_t.
\end{equation}
Here, we define $q_0 = q_{\text{data}}$, $g(t):=\sqrt{\frac{d\sigma^2(t)}{dt}}$, $*$ as the convolution operator, and $\bm{w}_t$ as the standard Wiener process. Eq.~\eqref{eq:sde_forward} inherently corresponds to a reverse time SDE from $T$ to $0$ under moderate conditions~\citep{anderson1982reverse}
\begin{equation}\label{eq:sde_backward}
    d\bm{x}(t) =  - g^2(t) \grad{\bm{x}}\log q_t(\bm{x}(t))  dt + g(t) d\bar{\bm{w}}_t,
\end{equation}
where $\bar{\bm{w}}_t$ is a standard Wiener process in reverse time, and $q_t(\bm{x})$ denotes the ground truth marginal density of $\bm{x}(t)$ following Eq.~\eqref{eq:sde_forward}. 

The stochastic process in Eq.~\eqref{eq:sde_backward} automatically associates with a deterministic process, known as the probability flow (PF) ODE. This PF ODE governs the evolution of samples without any diffusion term and guarantees that the trajectories of the samples maintain identical marginal probability densities as the forward SDE (Eq.~\eqref{eq:sde_forward}). The PF ODE is expressed as follows:

\begin{equation}\label{eq:prob_ode_gt}
    \frac{d\bm{x}}{dt} (t)=  -\frac{1}{2}g^2(t) \grad{\bm{x}} \log q_t (\bm{x}(t)).
\end{equation}

Since $\nabla_{\bm{x}} \log q_t(\bm{x}(t))$ is typically unattainable in Eqs.~\eqref{eq:sde_backward} and \eqref{eq:prob_ode_gt}, the denoising score matching (DSM) loss~\citep{vincent2011connection,song2020score} 
is commonly employed to approximate $\nabla_{\bm{x}}\log q_t(\bm{x})$ by using a time-conditional neural network $\bm{s}_{\bm{\theta}}=\bm{s}_{\bm{\theta}}(\bm{x}, t)$ over the time interval $[t_0, T]$, where $t_0\geq0$ is chosen to be sufficiently small in practice. 

By substituting $\grad{\bm{x}} \log q_t(\bm{x})$ with the learned $\bm{s}_{\bm{\theta}}$ in the reverse time SDE described in Eq.~\eqref{eq:sde_backward}, and in the PF ODE given by Eq.~\eqref{eq:fp_gt}, we obtain parametric counterparts of the reverse time SDE for a stochastic process and the PF PDE for a deterministic process, respectively. Consequently, we have the choice to generate samples by numerically solving either the parametric SDE or the parametric PF PDE in reverse, starting from an initial sample drawn from a predefined prior.

\begin{table*}[th!]
  \centering
  \caption{Comparison of existing consistency-type models.}
  \renewcommand{\arraystretch}{1.1}
  \resizebox{\textwidth}{!}{
  \begin{tabular}{lcccc}
    \bhline{0.8pt}
        \textbf{Models} & \textbf{Purpose} & \textbf{Trajectory} & \textbf{Object of Eq.} & \textbf{Approach}\\
        \bhline{0.8pt}
        
        CDM~\citep{daras2023consistent} & Sample quality & Backward SDE & Samples & DSM + Martingale regularizer\\
        CM~\citep{song2023consistency} & Sampling speed  & PF ODE & Samples & Specific NN structure + New training scheme\\
        FP-Diffusion~\citep{Lai2022ImprovingSD} & Likelihood  & Score FPE (a PDE) & Scores & DSM + Score FPE-regularizer\\
        \bhline{0.8pt}
  \end{tabular}
  }
  \label{tb:consistency_comp}
\end{table*}

\begin{figure*}[th]
    \centering
    \subfigure[Illustration of Def.~\ref{def:consist_sde_denoiser}. \jc{A consistent SDE-denoiser indicates that the SDE-denoiser prediction $\bm{h}(\bm{x}, t)$ (endpoint of the magenta arrow) aligns with the average of SDE predictions $\mathbb{E}_{p_{[t_0,t],\bm{h}}^{\text{SDE}}}\big[\bm{x}(t_0)|\bm{X}_t = \bm{x} \big]$ (blue dot). }]{
    \includegraphics[width=0.31\textwidth]{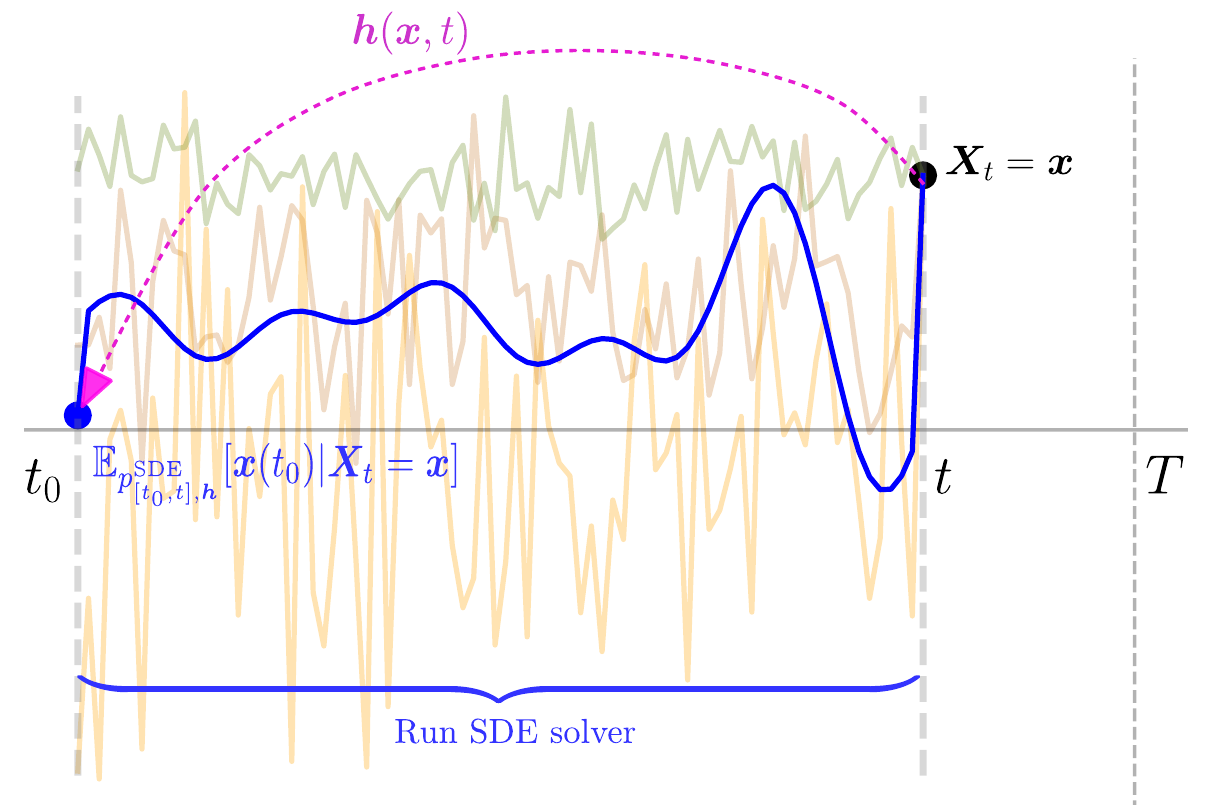}}
    \hspace{0.1cm}
    \subfigure[Illustration of Prop.~\ref{th:daras_equiv_martingale}. \jc{The SDE-denoiser prediction $\bm{h}(\bm{x}, t)$ (endpoint of the magenta arrow) aligns with the average prediction of intermediate points obtained by first applying an SDE solver and subsequently applying the SDE-denoiser $\mathbb{E}_{p_{[t',t],\bm{h}}^{\text{SDE}}}\big[\bm{h}(\bm{x}(t'), t')|\bm{X}_t = \bm{x} \big]$ (blue dot).}]{
    \includegraphics[width=0.31\textwidth]{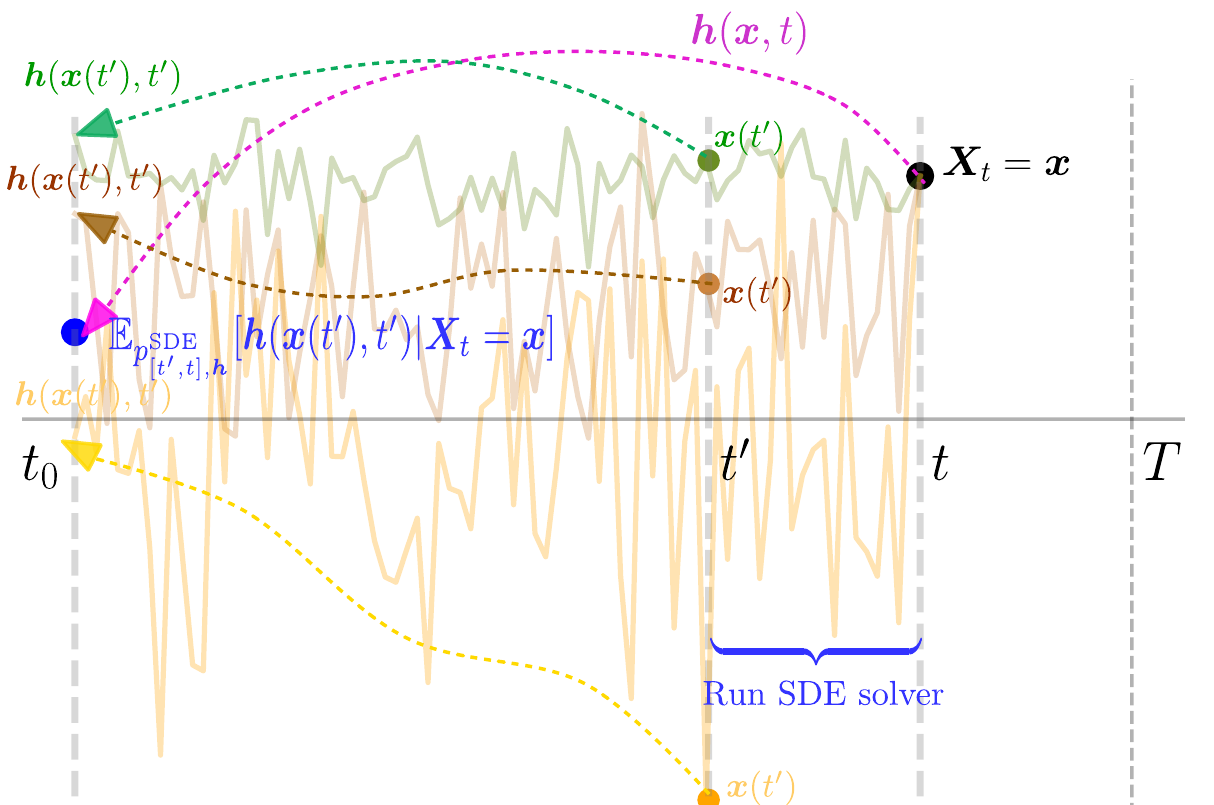}}
    \hspace{0.1cm}
    \subfigure[Illustration of Alg.~2 in \citep{song2023consistency}. \jc{The objective of CM is to align the prediction of the direct denoiser (endpoint of the magenta arrow) with the prediction obtained by first applying a one-step ODE solver and subsequently applying the denoiser.}]{
    \includegraphics[width=0.31\textwidth]{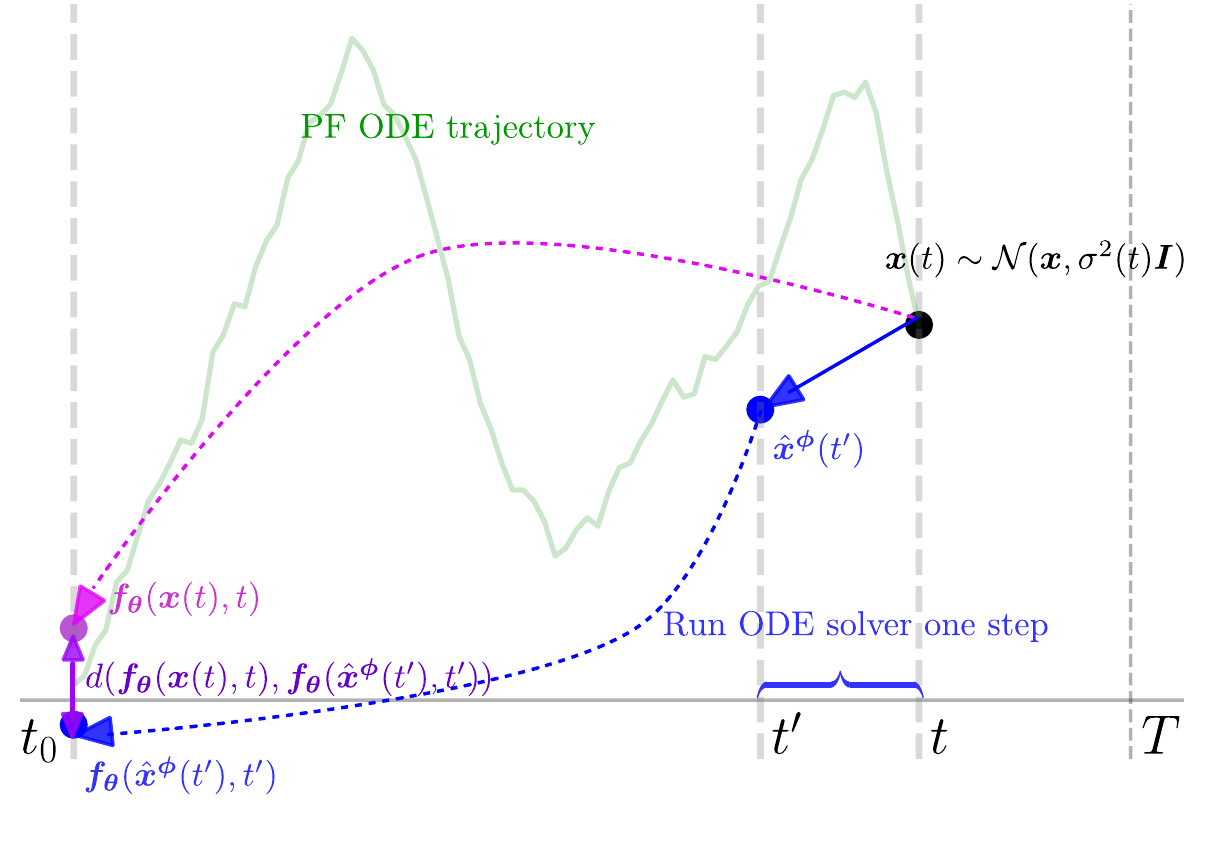}}
    \caption{Illustration of a consistent SDE/ODE-denoiser. The dashed curves depict predictions from denoisers.  In (a) and (b), the three erratic curves in light hues represent the SDE trajectories;  the blue curve in (a) indicates the deterministic trajectory of the average. In (c), the \jc{green} curve corresponds to the PF ODE trajectory, and $\bm{\phi}$ represents the parameters of a pre-trained score model. 
    }
    \label{fig:ill}
\vspace{-0.1cm}
\end{figure*}

\section{Consistency-type models}
\label{sec:consistency_type}

In this section, we provide an overview of recent literature that incorporates  notions of ``consistency'' in diffusion models. Specifically, we review three notable models: Consistent Diffusion Model (CDM)~\citep{daras2023consistent}, Consistency Model (CM)~\citep{song2023consistency}, and Fokker-Planck (FP) Diffusion~\citep{Lai2022ImprovingSD}. Table~\ref{tb:consistency_comp} compares the distinguishing characteristics of these consistency-type models.

\subsection{CDM~\citep{daras2023consistent}}
\label{subsec:cdm} 
\citet{daras2023consistent} introduced the concept of a ``consistent denoiser'' for the SDE~\eqref{eq:sde_backward}. By leveraging Tweedie's formula~\citep{efron2011tweedie}, a connection can be established between the score function $\nabla_{\bm{x}}\log q_t(\bm{x})$ and a denoiser $\bm{h}\colon\mathbb{R}^D\times[t_0, T]\rightarrow \mathbb{R}^D$ conditioned on time
\begin{equation*}
    \nabla_{\bm{x}}\log q_t(\bm{x}) = \frac{\bm{h}(\bm{x}, t)-\bm{x}}{\sigma^2(t)}.
\end{equation*}
Consequently, Eq.~\eqref{eq:sde_backward} can be rearranged as follows:
\begin{equation}\label{eq:sde_h_backward}
    d\bm{x}(t) =  - g^2(t) \Big(\frac{\bm{h}(\bm{x}, t)-\bm{x}}{\sigma^2(t)} \Big) dt + g(t) d\bar{\bm{w}}_t.
\end{equation}

 This reparameterization  gives rise to the concept of a consistent denoiser $\bm{h}$~\citep{daras2023consistent}. A denoiser is considered consistent if, on average, it produces estimates of the nearly clean data that align with those obtained by solving Eq.~\eqref{eq:sde_h_backward} in reverse, regardless of the initial data used. A concise summary of its formal definition is presented below. Furthermore, Fig.~\ref{fig:ill}(a) showcases its corresponding illustration.

\begin{definition}[Consistent SDE-denoiser~\citep{daras2023consistent}]\label{def:consist_sde_denoiser} A function $\bm{h}\colon\mathbb{R}^D\times[t_0, T]\rightarrow\mathbb{R}^D$ is called a \emph{consistent SDE-denoiser} if and only if $\bm{h}(\bm{x}, t) = \mathbb{E}_{p_{[t_0,t],\bm{h}}^{\text{SDE}}}\big[\bm{x}(t_0)|\bm{X}_t = \bm{x} \big]$  for all $\bm{x}\in\mathbb{R}^D$ and $t\in[t_0,T]$.

Here, $\mathbb{E}_{p_{[t_0,t],\bm{h}}^{\text{SDE}}}$ denotes the conditional expectation of $\bm{x}(t_0)$ with respect to the
distribution of $p_{[t_0,t],\bm{h}}^{\text{SDE}}$ along the stochastic trajectory described by the SDE presented in Eq.~\eqref{eq:sde_h_backward}. This trajectory starts with an initial value of $\bm{x}$ at an arbitrary time $t$ and terminates at a generated sample $\bm{x}(t_0)$ by running the SDE in Eq.~\eqref{eq:sde_h_backward} backwards in time.
\end{definition}

The aim of \citet{daras2023consistent} is to train a diffusion model to serve as a consistent SDE-denoiser. However, applying this condition to practical settings is challenging due to the time-consuming process outlined in Definition~\ref{def:consist_sde_denoiser}, which involves multiple SDE solving by running from $t$ to $t_0$ in order to accurately evaluate the average.
In contrast, \citet{daras2023consistent} observed that a consistent SDE-denoiser $\bm{h}$ can be interpreted as  a reverse martingale under the same process described in Eq.~\eqref{eq:sde_backward}. Fig.~\ref{fig:ill}(b) illustrates this property.

\begin{proposition}[\citet{daras2023consistent}]\label{th:daras_equiv_martingale} $\bm{h}$ is a consistent SDE-denoiser if and only if the following properties hold:
\begin{enumerate}
    \item[(i)] (Reverse martingale) For all $t>t'$ and $\bm{x}$, we have $\bm{h}(\bm{x}, t) = \mathbb{E}_{p_{[t',t],\bm{h}}^{\text{SDE}}}\big[\bm{h}(\bm{x}(t'), t')|\bm{X}_t = \bm{x} \big]$.
    \item[(ii)] (Identity at $t_0$) For all $\bm{x}\in\mathbb{R}^D$, $\bm{h}(\bm{x}, t_0)=\bm{x}$.
\end{enumerate}
Here, $\mathbb{E}_{p_{[t',t],\bm{h}}^{\text{SDE}}}$ represents the conditional expectation of $\bm{x}(t')$ given the distribution of $p_{[t',t],\bm{h}}^{\text{SDE}}$ along the trajectory described by the SDE in Eq.~\eqref{eq:sde_h_backward}, starting with an initial value $\bm{x}$ at time $t$ and terminates at time $t'$. 
\end{proposition}

Based on this proposition, \citet{daras2023consistent} proposed to train a denoiser $\bm{h}_{\bm{\theta}}$ by using the denoising score matching (DSM) loss, along with a regularizer which is defined as 
\begin{equation}\label{eq:cdm_reg}
    \frac{1}{2}\Big(\bm{h}_{\bm{\theta}}(\bm{x}, t) - \mathbb{E}_{p_{[t',t],\bm{h}_{\bm{\theta}}}^{\text{SDE}}}\big[\bm{h}_{\bm{\theta}}(\bm{x}(t'), t')|\bm{X}_t = \bm{x} \big]\Big)^2,    
\end{equation}
and its purpose is to enforce the reverse martingale property. Their approach demonstrates notable improvement in terms of sample quality.

\subsection{CM~\citep{song2023consistency}}
\label{subsec:cm}

\citet{song2023consistency} directed their attention to  PF ODE in Eq.~\eqref{eq:prob_ode_gt} (a deterministic process). They introduced the notion of a ``consistency function'' which promotes the model's prediction of nearly clean data that aligns with the trajectory of the PF ODE.
In order to establish a better connection with CDM (which will be discussed in Sec.~\ref{subsec:cdm_diff_cm}), we propose a modification to their terminology, replacing ``consistency function'' with ``consistent ODE-denoiser''. Below, we provide the formal definition.

\begin{definition}[Consistent ODE-denoiser~\citep{song2023consistency}]\label{def:consist_ode_denoiser} Given a solution trajectory $\{\bm{x}(t)\}_{t\in[t_0, T]}$ of the PF ODE in Eq.~\eqref{eq:prob_ode_gt}. 
Given a time-dependent vector field $\bm{f}$. $\bm{f}$ is called a  \emph{consistent ODE-denoiser} if it satisfies 
\begin{equation}\label{eq:consistent_ode_terminal}
    \bm{f}(\bm{x}(t),t)=\bm{x}(t_0)\quad \text{for all}\quad t\in[t_0, T].
\end{equation}
\end{definition}
A straightforward corollary of the definition is that a consistent ODE-denoiser $\bm{f}$ must fulfill the condition:
\begin{equation}\label{eq:consistent_ode_trajectory}
    \bm{f}(\bm{x}(t),t)=\bm{f}(\bm{x}(t'),t')\quad \text{for all}\quad t, t'\in[t_0, T].
\end{equation}

The goal of \citep{song2023consistency} is to train a network $\bm{f}_{\bm{\theta}}$ to satisfy both Eqs.~\eqref{eq:consistent_ode_terminal} and \eqref{eq:consistent_ode_trajectory}. Eq.~\eqref{eq:consistent_ode_terminal} is guaranteed by employing a specific network design, while Eq.~\eqref{eq:consistent_ode_trajectory} is learned through the minimization of a ``distance'' measured by $d(\cdot, \cdot)$:
 \begin{equation}\label{eq:cm_loss}
     d(\bm{f}_{\bm{\theta}}(\bm{x}(t),t), \bm{f}_{\bm{\theta}}(\bm{x}(t'),t')).
 \end{equation}
Thanks to the specific design of the network, the learned consistent ODE-denoiser is capable of performing one-step sampling. We visually depict the distillation training of \citet{song2023consistency} (Algorithm~2) in Fig.~\ref{fig:ill}(c).

\subsection{FP-Diffusion~\citep{Lai2022ImprovingSD}}
\label{subsec:fp_diff}
\citet{Lai2022ImprovingSD} established an equivalent system of PDEs known as the ``\emph{score Fokker-Planck equation}'' (\emph{score FPE}) for the ground truth score (see Eq.~\eqref{eq:fp_gt}), which is built upon the classic FPE introduced by \citet{fokker1914mittlere,planck1917satz}. The score FPE describes the temporal evolution of the ground truth score once the forward SDE is given. We present their findings in the following proposition.

\begin{proposition}[\citet{Lai2022ImprovingSD}]\label{th:fp}
The ground truth score $\bm{s}=\bm{s}(\bm{x}, t) :=\nabla_{\bm{x}} \log q_t(\bm{x})$  satisfies the score FPE
\begin{equation}\label{eq:fp_gt}
\begin{aligned}
    \partial_t \bm{s} &= \frac{1}{2}g^2(t)\grad{\bm{x}}\big( \div{\bm{x}}(\bm{s}) + \norm{\bm{s}}^2_2\big). 
\end{aligned}
\end{equation}
\end{proposition}
\citet{Lai2022ImprovingSD} noted that DSM pre-trained score-based diffusion models often do not conform to the underlying score FPE (which is satisfied by the ground truth score). To address this issue, they proposed a regularization approach by minimizing the residuals of the score FPE, referred to as the \emph{score FPE-regularizer}, into the DSM loss (as described in Eq.~(19) of \citet{Lai2022ImprovingSD}). The authors provided both theoretical and numerical evidence that their proposed model, known as \emph{FP-Diffusion}, can enhance likelihood estimation.

\section{Relation of consistency-type models}
\label{sec:relation}

\subsection{CDM and CM}
\label{subsec:cdm_diff_cm}
In this section, we establish a connection between CDM and CM. Specifically, we demonstrate that the notion of a consistent SDE-denoiser can be transformed into a consistent ODE-denoiser when the underlying trajectory is governed by a PF ODE.

To rigorously formulate the theorem, we incorporate a parameter $\lambda\geq 0$  that establishes a connection between Eqs.~\eqref{eq:sde_backward} and \eqref{eq:prob_ode_gt} as \citet{karras2022elucidating}
\begin{align}\label{eq:general_backward_sub}
    d\bm{x}(t) =  - (\frac{1+\lambda}{2})g^2(t) \bm{s}(\bm{x}, t)   dt + \lambda g(t) \bar{\bm{w}}_t.
\end{align}
Furthermore, we introduce $\mathbb{E}_{p_{[t',t],\bm{h}}^{\lambda-\text{SDE}}}$ analogous to Proposition~\ref{def:consist_sde_denoiser}, to denote the conditional expectation of $\bm{x}(t')$ based on the distribution of $p_{[t',t],\bm{h}}^{\lambda-\text{SDE}}$ along the stochastic trajectory described by the $\lambda$-parametrized SDE presented in Eq.~\eqref{eq:general_backward_sub} starting from $\bm{x}(t)$ and terminating at time $t'$. It is observed that when $\lambda=1$, Eq.~\eqref{eq:general_backward_sub} coincides with Eq.~\eqref{eq:sde_backward} and $p_{[t',t],\bm{h}}^{1-\text{SDE}}$ is equivalent to $p_{[t_0,t],\bm{h}}^{\text{SDE}}$. On the other hand, when $\lambda=0$, Eq.~\eqref{eq:general_backward_sub} becomes the PF ODE in  Eq.~\eqref{eq:prob_ode_gt}.  We now present the relationship of CDM and CM as the following theorem.

\begin{theorem}[Relationship of CDM and CM]\label{th:equiv_cdm_cm}  If $\lambda=0$, then a consistent SDE-denoiser is equivalent to a consistent ODE-denoiser.
\end{theorem}




\begin{proof}
    Consider a denoiser $\bm{h}$ and $\lambda=0$. This is equivalent to the formulation presented in Definition~\ref{def:consist_sde_denoiser}, but replacing the trajectory governed by the SDE in Eq.~\eqref{eq:sde_h_backward} with a PF ODE described by Eq.\eqref{eq:prob_ode_gt}. Then for all $t\in[t_0, T]$, we have
    $\bm{h}(\bm{x}(t), t) = \mathbb{E}_{p_{[t_0,t],\bm{h}}^{0-\text{SDE}}}\big[\bm{x}(t_0)|\bm{X}_t = \bm{x}(t) \big] = \bm{x}(t_0)$, owing to the deterministic nature of the ODE. Alternatively, we can arrive at the same conclusion by leveraging Proposition~\ref{th:daras_equiv_martingale}.  The reverse martingale property implies that $\bm{h}(\bm{x}(t), t) = \mathbb{E}_{p_{[t',t],\bm{h}}^{0-\text{SDE}}}\big[\bm{h}(\bm{x}(t'), t')|\bm{X}_t = \bm{x}(t) \big] = \bm{h}(\bm{x}(t'), t')$ holds for all $t'<t$, which is a consequence of the ODE trajectory being deterministic. In particular, if we set $t'=t_0$ and consider condition (ii) of Proposition~\ref{th:daras_equiv_martingale}, we also obtain $\bm{h}(\bm{x}(t), t) = \bm{h}(\bm{x}(t_0), t_0) = \bm{x}(t_0)$ for all $t\in[t_0, T]$. 
\end{proof}


\jc{Both perspectives presented in the proof, namely Definition~\ref{def:consist_sde_denoiser} and Proposition~\ref{th:daras_equiv_martingale}, ultimately yield the same conclusion: the theoretical correlation between Definition~\ref{def:consist_sde_denoiser} and Definition~\ref{def:consist_ode_denoiser}.} Indeed, Figs.~\ref{fig:ill}(b) and (c) illustrate an \jc{analogy} between a consistent SDE-denoiser and ODE-denoiser. Furthermore, it is worth noting that when we incorporate the CDM's regularizer in Eq.~\eqref{eq:cdm_reg} along the PF ODE, it simplifies to $\frac{1}{2}\big(\bm{h}_{\bm{\theta}}(\bm{x}, t) - \bm{h}_{\bm{\theta}}(\bm{x}(t'), t')\big)^2$, which coincides with Eq.~\eqref{eq:cdm_reg} when the measurement $d$ is taken as a $\frac{1}{2}$-weighted MSE.



\subsection{FP-Diffusion and CDM}
\label{subsec:fp_diff_cdm}
 According to \citet{daras2023consistent},  a consistent SDE-denoiser $\bm{h}$ is sufficient to guarantee the fulfillment of the score FPE by its corresponding induced score $\bm{s}(\bm{x}, t):=\frac{\bm{h}(\bm{x}, t)-\bm{x}}{\sigma^2(t)}$. \jc{In this section, we establish the necessity of this condition, thereby demonstrating the equivalence between these two concepts.}
 
 



\begin{theorem}[Equivalence of consistent SDE-denoiser and satisfaction of score FPE]\label{th:equiv_denoiser_scorefpe}
Let $\bm{h}\colon\mathbb{R}^D\times[t_0,T]\rightarrow\colon\mathbb{R}^D$ be a smooth denoiser so that $\bm{h}(\bm{x}, t_0) =\bm{x}$, for all $\bm{x}$.
Then $\bm{h}$ is a consistent SDE-denoiser if and only if $\bm{s}$ satisfies Eq.~\eqref{eq:fp_gt}. 
    
\end{theorem}
\begin{proof} The proof is built upon the approach presented in \citep{daras2023consistent}. It is important to note that once the forward SDE (Eq.~\eqref{eq:sde_forward}) is defined,  it naturally corresponds to a reverse SDE (Eq.~\eqref{eq:sde_backward}). By employing the multidimensional It\^o's Lemma~\citep[Lemma~A.2]{daras2023consistent} to Eq.~\eqref{eq:sde_backward}, we obtain
\begin{equation}\label{eq:ito_reverse}
    d\bm{h} =\Big[ \frac{\partial \bm{h}}{\partial t} -\frac{1}{2}g^2(t)\Delta_{\bm{x}}\bm{h} - g^2(t)\mathcal{J}_{\bm{h}}\cdot\bm{s} \Big] dt + g(t)  \mathcal{J}_{\bm{h}}d\bar{\bm{w}}_t.
\end{equation} 
 Here $\big(\Delta_{\bm{x}}\bm{F}(\bm{x}, t)\big)_i:=\sum_{j=1}^D\partial^2_{x_j}F_i(\bm{x},t)$ denotes the $\bm{x}$-Laplacian  of a vector field $\bm{F}:=(F_i)_{i=1}^{D}\colon\mathbb{R}^D\times[t_0, T]\rightarrow\mathbb{R}^D$ with a fixed time $t$. Moreover, $\mathcal{J}_{\bm{F}}$ denotes the Jacobian of $\bm{F}$ with respect to $\bm{x}$  (with a fixed time $t$).
We know from Proposition~\ref{th:daras_equiv_martingale} of \citet{daras2023consistent} that $\bm{h}$ is a consistent SDE-denoiser if and only if $\bm{h}$ is a reverse martingale. Indeed, it is equivalent to driftlessness of Eq.~\eqref{eq:ito_reverse} by applying Proposition~\ref{th:martingale_driftless}. Namely, the following equation is valid
\begin{equation*}
    \partial_t\bm{h} = \frac{1}{2}g^2(t)\Delta_{\bm{x}}\bm{h} + g^2(t)\mathcal{J}_{\bm{h}}\cdot\bm{s}.
\end{equation*}
 Finally, Lemma~\ref{th:equiv_s_h}, establishing a connection between the score FPE and a PDE that a denoiser $\bm{h}$ must satisfy, implies the equivalence of fulfilling the score FPE by $\bm{s}$.
\end{proof}


\jc{
Despite the theoretical equivalence of a consistent SDE-denoiser and the fulfillment of its score FPE, the empirical experiments conducted by \citet{daras2023consistent} and \citet{Lai2022ImprovingSD} used different approaches and achieved different outcomes. \citet{daras2023consistent} enforced consistency through regularization to promote the martingale property, whereas \citet{Lai2022ImprovingSD} applied a regularizer to ensure satisfaction of the score FPE. These different  methods result in distinct loss landscapes and optimization dynamics. We defer the empirical investigation of consistency-type models to future research.}


\section{Conclusion}
\label{sec:conclusion}
In this research, we provide a theoretical bridge between different consistency-type models. The results of this study have the potential to inspire the development of a comprehensive framework that ensures consistency and facilitates the simultaneous achievement of several desired benefits, such as accurate likelihood estimation, efficient sampling speed, and high sample quality.

\clearpage
\newpage
\bibliography{sbm_bib}
\bibliographystyle{icml2023}

\newpage
\appendix
\onecolumn

\section{Auxiliary lemmas and propositions}
\label{sec:aux_lemmas_props}

In this section, we present essential lemmas for proving Theorem~\ref{th:equiv_denoiser_scorefpe}.

Let us begin by revisiting the definition of a \emph{local martingale}~\citep{karatzas1991brownian}, which presents a localized version of the martingale property. A stochastic process $\{\bm{X}(t)\}_{t\in[t_0, T]}$ is considered a local martingale if there exists a sequence of stopping times $\{\tau_k\}_{k=1}^{\infty}$ (which are random variables) satisfying the following conditions: (i) $\tau_k$ is almost surely increasing, (ii) $\tau_k$ diverges to $\infty$ almost surely, and (iii) the stopped process $\bm{X}(\min\{t, \tau_k\})$ is a martingale.

It is well-known that a diffusion process without drift (e.g., Eq.~\eqref{eq:ito_form_sde}) is a local martingale in general. However, it is important to note that without additional conditions, it does not necessarily qualify as a true martingale. Here, we demonstrate a sufficient condition to imply the true martingale property of a local martingale. For the proof of this proposition, we refer to the provided source \citep{karatzas1991brownian}. 
\begin{lemma}\label{th:suff_martingale}  Suppose that $\mathbb{E}\big[\int_{t_0}^T \norm{\bm{G}(\bm{X}(\tau),\tau)}_2^2d\tau\big]<\infty$. Then the It\^o integral defined as 
    \begin{align}\label{eq:ito_form_sde}
        \bm{X}(t) = \bm{X}(t_0) +\int_{t_0}^{t} \bm{G}(\bm{X}(\tau),\tau)d\bm{w}_{\tau}
    \end{align}
is a martingale.    
\end{lemma}

Next, we present a classic result that establishes the constancy of any continuous local martingale with bounded variation. The proof of this lemma can be found in the reference \citep{karatzas1991brownian,schilling2021brownian}.
\begin{lemma}\label{th:bdd_v_to_constant} 
Any continuous local martingale of bounded variation is constant.
\end{lemma}

Now, we can establish a characterization of the martingale property for solutions of SDEs, which asserts the non-trivial equivalence between a martingale and a ``driftless'' It\^o process.

\begin{proposition}\label{th:martingale_driftless} 
     Assume that a (It\^o's) stochastic process $\{\bm{X}(t)\}_{t\in[t_0,T]}$ is the strong solution of the following reverse time SDE on $[t_0,T]$
    \begin{equation}\label{eq:sde_general}
        d\bm{X}(t) = \bm{F}(\bm{X}(t), t)dt+\bm{G}(\bm{X}(t), t)d\bar{\bm{w}}_t.
    \end{equation}
Here $\bm{F}, \bm{G}\colon\mathbb{R}^D\times[t_0,T]\rightarrow\mathbb{R}^D$ are vector fields satisfying some smoothness conditions~\footnote{Please refer to \citep[Theorem 5.2.1]{oksendal2003stochastic}.}, and assumed to be Lipschitz and sub-linear. Namely, there is a constant $C>0$ such that  
\begin{enumerate}
    \item[(i)] (Lipschitzness) For all $\bm{x}$, $\bm{y}\in\mathbb{R}^D$ and $t\in[t_0,T]$  \begin{align*}
        \norm{\bm{F}(\bm{x},t)-\bm{F}(\bm{y},t)}_2 &+ \norm{\bm{G}(\bm{x},t)-\bm{G}(\bm{y},t)}_2 \leq C \norm{\bm{x} - \bm{y}}_2.
    \end{align*}
    \item[(ii)] (Sub-linearity) For all $\bm{x}\in\mathbb{R}^D$ and $t\in[t_0,T]$
    \begin{align*}
        \norm{\bm{F}(\bm{x}, t)}+ \norm{\bm{G}(\bm{x},t)}\leq C (1+\norm{\bm{x}}_2).
    \end{align*}
\end{enumerate}
    Then $\bm{X}(t)$ is a reverse martingale if and only if Eq.~\eqref{eq:sde_general} is driftless, i.e., $\bm{F}\equiv 0 $.
\end{proposition}
The proposition can be understood intuitively by taking expectation of Eq.~\eqref{eq:sde_general} conditioned on its history $\{\bm{X}(s)\}_{s\leq t}$ 
\begin{equation*}
    \frac{d\mathbb{E}[\bm{X}(t)]}{dt} = \mathbb{E}[\bm{F}(\bm{X}(t),t)].
\end{equation*}
$\mathbb{E}[\bm{G}(\bm{X}(t),t)d\bm{w}_t]$ is zero because of the zero mean of the standard Wiener process $\bm{w}_t$. Therefore, the expectation remains constant (and consequently, independent of the past) if and only if $\bm{F}$ is identically equal to zero. Nevertheless, proving this proposition requires more intricate analysis, and we provide a detailed argument in Appx.~\ref{subsec:martingale_driftless}.


In the next Lemma, we bridge the score FPE and a PDE that a denoiser $\bm{h}$ should satisfy. To be consistent with notations in \citep{daras2023consistent}, we can derive the score FPE (Eq.~\eqref{eq:fp_gt}) of the ground truth score $\bm{s}$ as
 \begin{equation}\label{eq:score_equiv_pde}
     \partial_t\bm{s} = \frac{1}{2}g^2(t)\Delta_{\bm{x}}\bm{s} + g^2(t)\mathcal{J}_{\bm{s}}\cdot\bm{s}.
 \end{equation}
Here $\big(\Delta_{\bm{x}}\bm{F}(\bm{x}, t)\big)_i:=\sum_{j=1}^D\partial^2_{x_j}F_i(\bm{x},t)$ denotes the $\bm{x}$-Laplacian  of a vector field $\bm{F}:=(F_i)_{i=1}^{D}\colon\mathbb{R}^D\times[t_0, T]\rightarrow\mathbb{R}^D$ with a fixed time $t$. Moreover, $\mathcal{J}_{\bm{F}}$ denotes the Jacobian in $\bm{x}$ of $\bm{F}$ (with a fixed time $t$).
 
\begin{lemma}\label{th:equiv_s_h}
 $\bm{s}$ satisfies the score FPE (or equivalently, Eq.~\eqref{eq:score_equiv_pde}) if and only if  $\bm{h}$ satisfies
 \begin{equation}\label{eq:h_pde}
     \partial_t\bm{h} = \frac{1}{2}g^2(t)\Delta_{\bm{x}}\bm{h} + g^2(t)\mathcal{J}_{\bm{h}}\cdot\bm{s}.
 \end{equation}
\end{lemma}



\section{Proofs}
\subsection{Proof of Proposition.~\ref{th:martingale_driftless}}
\label{subsec:martingale_driftless}
\begin{proof} We establish the case for forward time SDEs and martingales, as the reverse time case can be derived through a similar line of reasoning~\citep{klenke2013probability}.  We first prove the sufficient implication. Suppose that $\bm{X}(t)$ is a martingale. Then the It\^o process $\bm{U}(t)$ defined as  
\begin{align*}
    \bm{U}(t):=&\int_{t_0}^{t} \bm{F}(\bm{X}(\tau),\tau)d\tau
    \\=&~\bm{X}(t) - \bm{X}(t_0) -\int_{t_0}^{t} \bm{G}(\bm{X}(\tau),\tau)d\bm{w}_{\tau}
\end{align*}
is a local martingale by the preservation of the local martingale property. It can be observed that $\bm{U}(t)$ possesses a continuous path and bounded variation on the interval $[t_0, T]$, as a result of property (i) and the assumed smoothness. By utilizing Lemma~\ref{th:bdd_v_to_constant}, we conclude that $U(t)$ is constant, specifically zero. Consequently, Eq.~\eqref{eq:sde_general} is driftless.

Next, we prove the necessity. Suppose that $\bm{F}\equiv 0 $. Thanks to conditions (i) and (ii) (and some additional technical conditions), the existence and uniqueness of the solution to Eq.~\eqref{eq:sde_general} is guaranteed (see in \citep[Theorem 5.2.1]{oksendal2003stochastic}) and the solution is finite in the $L_2$-sense
\begin{align*}
    \mathbb{E}\big[\int_{t_0}^{T} \norm{X(\tau)}_2^2d\tau \big]<\infty.
\end{align*}
We will show that it implies $\mathbb{E}\big[\int_{t_0}^T \norm{\bm{G}(\bm{X}(\tau),\tau)}_2^2d\tau\big]<\infty$. Consequently, according to Lemma~\ref{th:suff_martingale}, it guarantees that $\bm{X}$ is a martingale. The sub-linearity of $\bm{G}$ indicates
\begin{align*}
    \norm{\bm{G}(\bm{x},t)}_2^2 \leq C^2(1+\norm{x}_2)^2 \leq 2C^2(1+\norm{x}_2^2).
\end{align*}
So 
\begin{align*}
\mathbb{E}\big[\int_{t_0}^T \norm{\bm{G}(\bm{X}(\tau),\tau)}_2^2d\tau\big] \lesssim(T-t_0)+\mathbb{E}\big[\int_{t_0}^T\norm{\bm{X}(\tau)}_2^2d\tau\big]<\infty.
\end{align*}
Here we use $\lesssim$ to absorb multiplicative constants. Consequently, utilizing Lemma~\ref{th:suff_martingale} and relying on the uniqueness of the strong solution, it follows that $\bm{X}$ is a martingale.
\end{proof}

\subsection{Proof of Lemma.~\ref{th:equiv_s_h}}

\begin{proof} Suppose that $\bm{s}$ satisfies Eq.~\eqref{eq:score_equiv_pde}.
    We know that $\bm{h}(\bm{x}, t)= \bm{x} + \sigma^2(t) \bm{s}(\bm{x}, t)$. By taking $\partial_t$ and using the chain rule, we obtain
    \begin{align}\label{eq:h_to_s}
        \begin{aligned}
            \partial_t\bm{h} &= \frac{d\sigma^2(t)}{dt}\bm{s}+\sigma^2(t) \partial_t\bm{s} \\
            &= g^2(t) \bm{s} + \sigma^2(t) \Big(\frac{1}{2}g^2(t)\Delta_{\bm{x}}\bm{s} + g^2(t)\mathcal{J}_{\bm{s}}\cdot\bm{s}\Big).
        \end{aligned}
    \end{align}
Notice that $\Delta_{\bm{x}}\bm{s}(\bm{x}, t) = \Delta_{\bm{x}}\big(\frac{\bm{h}(\bm{x}, t)-\bm{x}}{\sigma^2(t)}\big)=\frac{1}{\sigma^2(t)}\Delta_{\bm{x}}\bm{h}(\bm{x}, t)$ and that $\mathcal{J}_{\bm{s}}  = \frac{1}{\sigma^2(t)}(\mathcal{J}_{\bm{h}} - \bm{I})$, where $\bm{I}$ denote the identity matrix on $\mathbb{R}^D$. So Eq.~\eqref{eq:h_to_s} becomes
    \begin{align*}
        \begin{aligned}
            \partial_t\bm{h} &= 
             g^2(t) \bm{s} + \frac{1}{2}g^2(t)\Delta_{\bm{x}}\bm{h} + g^2(t)(\mathcal{J}_{\bm{h}} - \bm{I})\cdot\bm{s}
             \\&=\frac{1}{2}g^2(t)\Delta_{\bm{x}}\bm{h} + g^2(t)\mathcal{J}_{\bm{h}}\cdot\bm{s},
        \end{aligned}
    \end{align*}
which indicates Eq.~\eqref{eq:h_pde} is fulfilled by $\bm{h}$. The reverse implication follows with a similar computation. 
\end{proof}


\end{document}

%% file: math_commands_icml.tex
\usepackage{graphicx}
\usepackage{wrapfig}
\usepackage[para,online,flushleft]{threeparttable}
\usepackage{algorithm,algorithmic}
\usepackage{hyperref}
\usepackage{xurl}
\usepackage{verbatim}
\usepackage{varwidth}
\usepackage{threeparttable}
\usepackage{tabularx}
\usepackage{multirow}
\usepackage{amsmath,amsfonts,bm,amssymb,amsthm}
\usepackage{mathrsfs}  
\usepackage{array,arydshln}
\usepackage{threeparttable}
\usepackage{graphicx}
\usepackage[super]{nth}
\usepackage{booktabs}
\usepackage{xcolor}
\usepackage[normalem]{ulem}
\usepackage{caption}
\usepackage{lipsum}
\newcommand{\bhline}[1]{\noalign{\hrule height #1}}

\newcommand{\grad}[1]{{\nabla_{\bm{#1}}}}
\renewcommand{\div}[1]{{\textup{div}_{\bm{#1}}}}

\def\eqref#1{(\ref{#1})}

\def\1{\bm{1}}

\DeclareMathAlphabet{\mathsfit}{\encodingdefault}{\sfdefault}{m}{sl}
\SetMathAlphabet{\mathsfit}{bold}{\encodingdefault}{\sfdefault}{bx}{n}

\newcommand{\norm}[1]{\left\lVert#1\right\rVert}

\definecolor{ube}{rgb}{0.53, 0.47, 0.76}
\definecolor{ballblue}{rgb}{0.13, 0.67, 0.8}

\newcommand{\jc}[1]{{\color{black}{{#1}}}}

\definecolor{spink}{rgb}{0.831,0.184,0.494}